# A Hierarchical Probabilistic Model for Facial Feature Detection

Yue Wu     Ziheng Wang     Qiang Ji
ECSE Department, Rensselaer Polytechnic Institute
{wuy9,wangz10,jiq}@rpi.edu

## Abstract

*Facial feature detection from facial images has attracted great attention in the field of computer vision. It is a nontrivial task since the appearance and shape of the face tend to change under different conditions. In this paper, we propose a hierarchical probabilistic model that could infer the true locations of facial features given the image measurements even if the face is with significant facial expression and pose. The hierarchical model implicitly captures the lower level shape variations of facial components using the mixture model. Furthermore, in the higher level, it also learns the joint relationship among facial components, the facial expression, and the pose information through automatic structure learning and parameter estimation of the probabilistic model. Experimental results on benchmark databases demonstrate the effectiveness of the proposed hierarchical probabilistic model.*

## 1. Introduction

Accurately analyzing the face is critical, since the face reflects the psychological and physical states of humans. For example, by analyzing the face, we can identify human facial expressions [8], detect the fatigue state of human, estimate the human head pose [19], and so on. As the bottom level representation, the locations of facial feature points such as eyebrow tips, eye corners, mouth corners provide a promising way to describe the face shape and capture the rigid and nonrigid deformation of the face.

Detection of these facial feature points is challenging, since facial appearance and shape tend to change significantly under different facial expressions and poses. For example, in Figure 1, the local appearance and shape of the mouth varies dramatically. In (a), the lip corners pull up. The mouth opens widely in (b), while closes tightly in (c). Therefore, a good facial feature detection method should be able to model the shape variations under different situations. Another important observation is that facial components do not change independently and there exists interaction among facial components. For example, in (b), eye lids go up and the mouth opens widely simultaneously. As a result, it is also important to model the relationship among facial components. Furthermore, since facial expression and pose are two major causes of the significant change of the face shape, it is necessary to take these information into account when modeling the face shape variations.

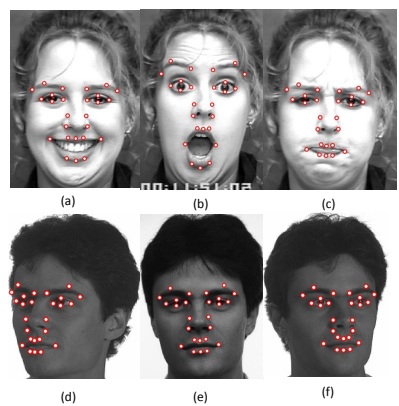

Figure 1. Facial features detection for facial images with facial expressions and poses.

Following these three observations, in this paper, we propose a hierarchical probabilistic model to capture the the variations of face shapes under significant facial expression and pose changes, and infer the true facial feature locations given the measurements. Specifically, we first model the local shape variations of each facial component using a mixture model. In addition, we model the joint interactions of facial components in the higher level. Unlike most of the previous methods, we explicitly add the facial expression and pose information to help construct and parameterize the model. During detection, the true facial feature locations are inferred through the bottom-up lower level shape constraints of facial components and the top-down constraint from the relationship among facial components.

The remainder of this paper is organized as follows: Section 2 reviews the related work. In section 3, we will describe the model, its learning and inference. Section 4 will show the experimental results on several benchmark databases. Section 5 will conclude the paper.

## 2. Related Work

As a promising way to describe the shape of the nonrigid face, facial feature localization has gained much more attention in computer vision community over the past years. Methods for facial feature detection can be classified into three categories: holistic methods, local methods and component based methods.

**Holistic methods:** The holistic methods build models to capture the appearance and shape of the whole facial image. Typical methods are the classic Active Shape Model (ASM) [2] and Active Appearance Model (AAM) [1]. ASM is a statistical model that uses the mean shape and eigenvectors generated from PCA to represent the deformations of objects' shape. AAM uses both the texture and shape information of the whole facial image to fit a linear generative model. However, it is sensitive to the lighting conditions and performs poorly on unseen identities [11].

**Local methods:** The local methods use local image features. They build a likelihood map for each landmark based on the response of a regression function or a classifier. Usually, these likelihood maps are combined with a global shape model as a constraint. In [3], Cristinacce and Cootes proposed the Constrained Local Model (CLM). It generates the likelihood map for each facial feature using the matching score of the local testing image to the templates. Then, the likelihood maps are combined with a prior distribution of the shape that is similar to ASM. More recently, in [25], Valstar et al. proposed the BoRMaN model that combines Boosted Regression with Markov Networks. It generates the likelihood map for each point based on the response of the support vector regressor, and then constraints the point locations by the shape distribution embedded in a pre-trained Markov Random Fields model. In [18], Martinez et al. proposed a Local Evidence Aggregation for Regression (LEAR) based facial feature detection method. It utilizes similar shape constraint as [25] with improved searching procedure that considers the aggregation of local evidence. In [22], instead of using the parametric model, the work of Saragih et al. represents the face shape prior model in a non-parametric manner with optimization strategy to fit facial images.

**Component based methods:** The component based methods focus on molding the shape variations of each facial component. The relationships among different facial components are further embedded in the model as a global constraint. In [12], a bi-stage model was proposed to characterize the face shape locally and globally. The local shape variation is modeled as a single Gaussian, and the nonlinear relationship among components is represented as a pre-trained Gaussian process latent variable model. Similarly, [16] uses one PCA for local component shape and a global PCA to model locations of components. Tong et al. [24] proposed a model to capture the different states of facial components like mouth open and mouth closed. But, during testing, they need to dynamically and explicitly estimate the state of local components and switch between different states. In [7], for each facial component, Ding and Martinez emphasized distinguishing the image patch centered at the true facial component location from the contextual patches with the use of the subclass division method.

Among all the facial feature localization methods, only a few works explicitly model the face shape variations under expression and pose changes. In [4], instead of constructing one single set of regression forests, Dantone et al. built sets of regression forests conditional on different poses. In [27], Wu et al. built a prior model for face shapes under varying facial expressions and poses. They decompose the variations into the parts caused by facial expressions and by poses using a 3-way Restricted Boltzmann Machine(RBM) model. However, it does not explicitly use the label information of facial expressions and poses during training. In [29], Zhu and Ramanan proposed the FPLL model that simultaneously performs face detection, pose estimation, and landmark localization. It builds a shape prior model based on the tree structure graphical model for each face pose.

In this paper, our major contribution is a hierarchical probabilistic model that could infer the true facial feature locations given image measurements. However, unlike some other component-based methods, we implicitly model the shape of facial components by learning the variations under different hidden states which is usually assigned or ignored by previous work. Furthermore, we explicitly learn the relationships among facial components, facial expression and poses, and use these relationships to improve facial feature detection accuracy.

## 3. Our Approach

### 3.1. Problem formulation

Given the initial locations of 26 facial feature points $X_m = [x_{m,1}, x_{m,2}, ..., x_{m,26}]$ generated using individual facial feature detectors, our goal is to estimate the true facial feature locations $X = [x_1, x_2, ..., x_{26}]$. Figure 1 shows the 26 facial feature points. We can formulate this task as an optimization problem under the probabilistic framework:

$$X^* = \arg\max_X P(X|X_m) \qquad (1)$$

Therefore, if we can construct a model to represent the relationship between $X_m$ and $X$, we can find the true facial feature locations by maximizing the posterior probability of $X$ given $X_m$. Towards this goal, we propose the following hierarchical probabilistic model.

## 3.2. Hierarchical probabilistic model

As shown in Figure 2 (b), our hierarchical model consists of nodes from four layers. Nodes in the lowest layer indicate the measurements of facial feature locations for each facial component. Nodes in the second layer represent the true facial feature locations we want to infer. In the third layer, all nodes are discrete latent random variables indicating the states of each facial component. The top layer contains two discrete nodes representing the facial expression and the pose.

This model captures two levels of information, including local shape variations of each facial component and the joint relationship among facial components, facial expression and pose. The lower level information is embedded among the nodes belonging to the same facial component. The higher level information is represented by the joint relationship among nodes within the rectangle, which are learned through training data. In the following subsections, we will first illustrate the model that captures this two level information, and its properties and benefits. Furthermore, we discuss how to generate the measurements.

### 3.2.1 Modeling the shape variations for each facial component

In order to capture the face shape variations more accurately, we follow the component based method and model the face shape variations of each facial component including the eyebrow, eye, nose and mouth. However, for each component, the shape may change significantly due to varying facial expressions and poses. So, to represent the shape variations with different conditions, we use the mixture model shown in Figure 2 (a). Here, $Z$ is a discrete variable that represents the hidden state of each facial component. $X$ indicates the true facial feature locations for this component, and $X_m$ is the measurement. For one component shown in Figure 2 (a), we model the joint probability of $X$ and $X_m$ as follows:

$$P(X, X_m) = \sum_Z P(X_m|X,Z)P(X|Z)P(Z) \quad (2)$$

$P(X|z)$ describes the face shape pattern under a specific hidden state $Z = z$. We can use a Multivariate Gaussian Distribution to model this probability.

$$P(X|z) \sim N(\mu_z, \sigma_z) \quad (3)$$

,where $\mu_z$ is the mean vector and $\sigma_z$ is the covariance matrix. $P(X_m|X, z)$ captures the relationship between measurement and inferred true facial feature locations for a specific state $z$. It can be modeled as a linear Multivariate Gaussian Distribution.

$$P(X_m|X, z) \sim N(\beta_{0,z} + \beta_z X, Q_z) \quad (4)$$

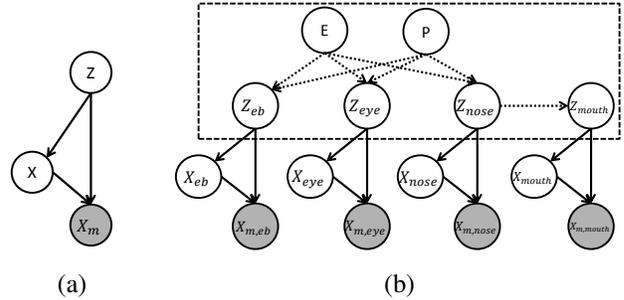

(a)          (b)

Figure 2. Hierarchical Probabilistic Model. (a) Component model. (b) A sample structure. Node connections marked as solid lines are fixed while these with dotted lines are learned.

, where $\beta_{0,z} + \beta_z X$ is the mean of the Gaussian and $Q_z$ is the covariance matrix. $P(Z)$ is the probability for different hidden states.

Figure 3 shows some examples of the shape variations of mouth under different hidden states that are learned from data with expression variations. It can be seen that, different hidden states relate to different facial expressions. The combinations of all states describe the overall shape variations of mouth due to facial expression changes.

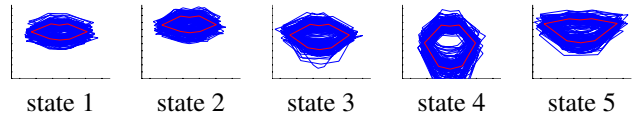

state 1     state 2     state 3     state 4     state 5

Figure 3. Examples of the shape variations of the mouth for different hidden states. Learned from data with expression variations.

### 3.2.2 Modeling the relationship among facial components, facial expression and pose

The states of different facial components are highly related. For example, if people lower their eyebrows, it's difficult to open their eyes widely simultaneously. So, it's important to take advantage of relationships among facial components. Facial expression and pose are two major causes of the significant change of the face shape. As a result, by explicitly adding the label of facial expression and pose for all the training data, the model will be more accurate than one without this additional information.

The relationships among facial components, facial expression and pose are represented as the joint probability of all the nodes inside the dotted rectangle. It is important to notice that all the $Z_i$ nodes are hidden and the underlying relationship among $Z_i$, $E$ and $P$ can not be assigned arbitrarily. Consequently, we need to automatically learn the model structure within the rectangle and estimate the parameters of the entire model. The dotted lines within the rectangle in Figure 2 (b) are shown as an special example for illustration purposes only.

With the combination of local shape models for each facial component, and the global relationship among facial components, facial expression and pose, the overall joint probability of $X$ and $X_m$ shown in Figure 2 (b) is formulated as:

$$P(X, X_m) = \sum_{Z,E,P} (\prod_{i \in \{eb,e,n,m\}} P(X_{m,i}, X_i|Z_i))P(Z, E, P), \quad (5)$$

where $Z = [Z_{eb}, Z_e, Z_n, Z_m]$.

### 3.2.3 Properties of the proposed model

There are a few important properties and benefits of the proposed hierarchical probabilistic model.

1. The local shape variation of each facial component is modeled implicitly with hidden states. As a result, it could more accurately capture the variations under different conditions. Furthermore, unlike [24] we do not need to assign any particular meaning for each state such as mouth open and mouth closed.

2. The relationship among facial components are represented by the joint distribution of all the hidden nodes $Z_i$. This relationship is automatically discovered based on bottom-up information from the facial feature locations $X_i$ and the top-down information from facial expression and pose which are available during training. In this case, the hidden states relate to the major causes of the shape variations, but they also can represent the information within the training data.

3. Facial expression and pose information are the major causes of the face shape variation. In this model, they explicitly help the learning of both the local shape variations and the joint relationship among facial components. This important information is usually ignored by the previous work.

### 3.2.4 Facial feature measurements

An integral part of our model is the facial feature measurements ($X_m$). They serve as input to the proposed hierarchical model, help learn the model during training, and are used to infer the true facial feature positions during testing. The facial feature measurements $X_m = [x_{m,1}, x_{m,2}, ..., x_{m,26}]$ are the initial locations of 26 facial feature points. To generate the measurements, we first detect eyes on images based on the Viola-Jones eye detector [26]. The detected eye positions provide normalization and initial locations of the facial features. For each feature point, multi-scale and multi-orientation Gabor-wavelets [5] are employed to generate the representation. The initial feature point positions are further refined by a fast Gabor wavelet jets matching method [9] that computes a displacement to move the initial facial feature positions to their final positions subject to certain shape constraints learned from the training data.

### 3.3. Model learning

Given the true feature location $X$, its measurement $X_m$, its facial expression label $E$, and pose label $P$ as training data, we could learn the model. We refer to model learning as learning the model structure and the model parameters. As shown in Figure 2(b), for structure learning, we only learn the global structure, denoted as $\mathbf{M}_G$ that connects the nodes within the rectangular block, and we fix the local model $\mathbf{M}_L$. For parameter learning, we estimate the parameters of the whole model $\Theta = [\Theta_G, \Theta_L]$. To tackle this learning task, we applied the Structure EM algorithm [10] and modified it for our application.

The complete description of our learning algorithm is shown in algorithm 1. Given the training data, in the initialization step, the hidden states $Z^0 = [Z_{eb}, Z_e, Z_n, Z_m]$ for each data are first inferred by independent clustering of each local facial component, where the number of clusters is exhaustively searched on the validation data set. With the initial estimation of hidden states, initial model structure and parameters are estimated based on the complete data. Similar to structure EM algorithm [10], we iteratively search for better parameters and structure with three major steps, denoted as 1, 2, 3.

In step 1, standard parameter EM algorithm is applied to update the parameters. It keeps the structure unchanged and searches for the parameters that maximize the expected BIC score, as shown in Equation 6.

$$\Theta^{n,l+1} = \arg\max_\Theta \mathbf{Q}_{BIC}(M^n, \Theta : M^n, \Theta^{n,l}) \quad (6)$$

Here, $\mathbf{Q}_{BIC}(.)$ denotes the expected BIC score and it is defined in equation 7.

$$\mathbf{Q}_{BIC}(M^n, \Theta : M^n, \Theta^{n,l})$$
$$= \mathbf{E}[log P(E, P, X, X_m, Z : M^n, \Theta) - \frac{log N}{2} Dim(M^n, \Theta)], \quad (7)$$

where $\mathbf{E}[.]$ denotes the expectation over the probabilities of the incomplete data under the current model as $P(Z|E, P, X, X_m : M^n, \Theta^{n,l})$, and $Z = \{Z_{eb}, Z_e, Z_n, Z_m\}$. $N$ denotes the number of training data, and $Dim(.)$ represents the model complexity in terms of the number of independent parameters. The parameters are updated iteratively to take advantage of the small computational cost of parameter EM compared to structure learning.

In step 2, we search for the new structure that maximizes the expected BIC score given the current model and parameters as shown in Equation 8.

$$\mathbf{M}_G^{n+1} = \arg\max_{\mathbf{M}_G} \mathbf{Q}_{BIC}(\mathbf{M}_G, \Theta_G : M^n, \Theta^{n,l}), \quad (8)$$

where $\mathbf{Q}_{BIC}(.)$ has similar definition as in Equation 7. Because of the nice decomposable property of BIC score, we only need to search for the global model, $\mathbf{M}_G$, that maximizes the expected BIC score belongs to the nodes within the rectangular block. Here, we applied the structure learning method in [6] which guarantees a global optimality with respect to the expected score function based on branch and bound algorithm. It is important to note that the expected BIC score is calculated over $P(\mathbf{Z}|X, X_m, E, P : \mathbf{M}^n, \Theta^{n,l})$ which is inferred with the whole model using the junction tree algorithm [15]. In step 3, the initial parameters for the next iteration for the new structure are estimated with Equation 9.

$$\Theta^{n+1,0} = \arg\max_{\Theta} \mathbf{Q}_{BIC}(\mathbf{M}^{n+1}, \Theta : \mathbf{M}^n, \Theta^{n,l}). \quad (9)$$

The updating stops when the number of iterations reaches the maximal value allowed, or the structure remains the same in two consecutive iterations.

---

**Algorithm 1**: Learning the hierarchical probabilistic model

**Data**: Training data $\{\mathbf{X}_j, \mathbf{X}_{m,j}, \mathbf{E}_j, \mathbf{P}_j\}_{j=1}^N$
**Result**: Model structure $\mathbf{M}_G$ and model parameters $\Theta = [\Theta_G, \Theta_L]$
**Initialize:** Initialize $Z^0 = [Z_{eb}, Z_e, Z_n, Z_m]$ by independent clustering. With initial $Z^0$, learn initial structure $\mathbf{M}_G^0$ and parameters $\Theta^{0,0}$ using the complete data.
**for** $n = 0, 1...$ *until convergence* **do**
  **for** $l = 0, 1...l_{max}$ *or convergence* **do**
    /* Update parameters          */
1   $\Theta^{n,l+1} = \arg\max_\Theta \mathbf{Q}_{BIC}(M^n, \Theta : M^n, \Theta^{n,l})$
  **end**
  /* Update global model structure.   */
2  $\mathbf{M}_G^{n+1} = \arg\max_{\mathbf{M}_G} \mathbf{Q}_{BIC}(\mathbf{M}_G, \Theta_G : \mathbf{M}^n, \Theta^{n,l})$.
  The expectation is taken over the possibilities of hidden nodes $Z$ given previous model and observation denoted as $P(\mathbf{Z}|X, X_m, E, P : \mathbf{M}^n, \Theta^{n,l})$.
  $\mathbf{M}^{n+1} = [\mathbf{M}_G^{n+1}, \mathbf{M}_L]$
  /* Initialize parameters for next iteration.         */
3  $\Theta^{n+1,0} = \arg\max_\Theta \mathbf{Q}_{BIC}(\mathbf{M}^{n+1}, \Theta : \mathbf{M}^n, \Theta^{n,l})$
**end**

---

### 3.4. Inference

Once the structure and parameters of the model have been estimated, we have learned the joint probability of the true facial feature locations $X$ and the measurement $X_m$. So, inference of $X$ can be performed by maximizing the posterior probability:

$$X^* = \arg\max_X P(X|X_m) = \arg\max_X P(X, X_m) \quad (10)$$

It is important to note that, during inference, we marginalize all the discrete latent states $Z_i$, unknown facial expression $E$ and pose $P$. Here we use the junction tree algorithm [15] to perform the inference.

## 4. Experimental Results

In this section, we test our facial feature detection method on several benchmark databases, and compare it to some existing works.

### 4.1. Databases

**Facial images with expression variations:** Facial expression could dramatically change the appearance and shape of the face. To test our model on images with facial expressions, we used the **CK+** [13][17] database and the **MMI** [20] database. In our experiments, for each database, we used sequences corresponding to 6 basic facial expressions, including anger, disgust, fear, happiness, sadness and surprised. For each sequence, we used the first frame, onset and apex frames. In total, CK+ contains 1339 images and MMI contains 584 images.

**Facial images with pose variations:** Pose is another cause for the significant changes of face shape. To verify our model under this condition, we test it on the **FERET** database [21] and the Annotated Facial Landmarks in the Wild database (**AFLW**) [14]. In our experiments, for FERET, we used 335 frontal, quarter left and right images and ignore images in other poses since not all facial features were observable in other poses. AFLW contains images collected from the website with large facial appearance and shape variations. In our experiments, we used 4226 images for which all the annotation of inner points are available. To build our hierarchical model, we discretized the pose based on the ground truth yaw (less than"$-10^o$", between "$-10^o$" and "$10^o$", larger than "$10^o$") and pitch angles (less than "$-15^o$", between "$-15^o$" and "$0^o$", larger than "$0^o$") provided by the database and ignored the roll angle since it can be excluded by in-plane rotation.

**Facial images with expression and pose variations:** Possible face shape variations could be due to both facial expressions and poses. In our experiments, we used the **ISL** multi-view facial expression database [23] to evaluate our method. The ISL multi-view database contains sequences of 8 subjects showing happy and surprised facial expressions under varying continuous face poses. We selected 460 images consisting of about every 5 frames from each sequence and some key frames with significant pose or facial expression changes. We discretized the continuous pose into three states, including the frontal, left and right. For each image in all databases, we manually labeled the facial feature locations as ground truth.

### 4.2. Implementation details

To evaluate the performance of the detection algorithm, we used the distance error metric:

$$Error_{i,j} = \frac{\|P_{i,j} - \hat{P}_{i,j}\|_2}{D_I(j)} \quad (11)$$

,where $D_I(j)$ is the interocular distance measured at frame $j$, $\hat{P}_{i,j}$ is the detected point $i$ at frame $j$ and $P_{i,j}$ is the manually labeled ground truth. For each database except ISL, we tested our method based on 10 fold cross-validation strategy. For ISL, to make up the limited training data, we used 46 fold cross-validation.

### 4.3. Algorithm evaluation

#### 4.3.1 Facial feature detection on CK+ database

The learned hierarchical model for CK+ is shown in Figure 6 (a). It can be seen that, only the eyebrow and mouth directly link to the expression node. Eye and nose are independent of expression given the states of mouth. The learned local shape variations of the mouth have been shown in Figure 3, and the shape variations of eyebrow are shown in Figure 4. The learned states are highly related to different facial expressions. Figure 5 (a) explicitly illustrates the relationship between hidden states of mouth and facial expressions, which corresponds to the probability $P(Z_{mouth}|E)$ marked as the red bold line in Figure 6 (a), connecting $E$ and $Z_{mouth}$. The shape variation of the mouth under state 4 is highly related to the surprised facial expression while state 5 is caused by the happy facial expression. Another important benefit here is that shape variations with the surprised facial expression are modeled with the first 4 states which leads to a more accurate model than using one single state for each facial expression. Figure 5 (b) plots the joint probability $P(Z_{eyebrow}, Z_{mouth})$, which indicates the relationship between hidden states of the mouth and eyebrow. State 1 of the mouth and state 1 of the eyebrow co-occur quite often. State 4 of the mouth and state 3 of the eyebrow usually happen together. The later situation is due to the surprised facial expression and the former is because of the neutral expression.

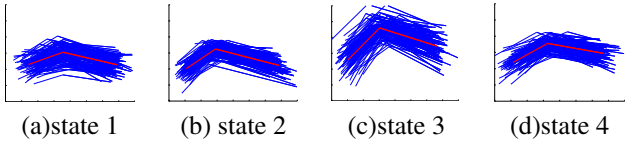

(a)state 1　　(b) state 2　　(c)state 3　　(d)state 4

Figure 4. Examples of the shape variations of eyebrow for different states. Learned on CK+.

The overall facial feature detection accuracy on the CK+ database is shown in Table 1. With the learned hierarchical model, performance is better than the manually constructed model without the hidden states shown in Figure 6 (b).

Table 1. Detection results using learned and manually constructed models on CK+ database.

|         | Eyebrow | Eye    | Nose   | Mouth  | Overall |
|---------|---------|--------|--------|--------|---------|
| Manual  | 5.5487  | 3.0357 | 5.3267 | 3.8645 | 4.2233  |
| Learned | **5.4287** | **2.8466** | **4.8856** | **3.8212** | **4.0560** |

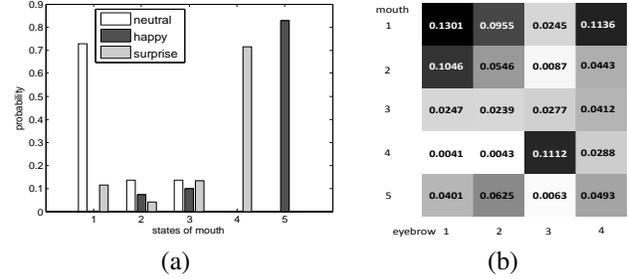

(a)　　　　　　　　　　(b)

Figure 5. (a)Probability of hidden states of mouth for "neutral", "happy" and "surprised" facial expression, denoted as $P(Z_{mouth}|E)$. (b) Co-occurrence matrix and joint probability of hidden states of mouth and eyebrow.

#### 4.3.2 Facial feature detection on other databases

Figure 6 (a)(c)(d) show some learned models based on data with different variations. Because of the different variations, top nodes in our hierarchial model could include expression, pose, or both nodes. The relationship among hidden states also varies, and it depends on the properties embedded within different training data. For example, the hidden states learned for data with pose variations on the FERET database shown in Figure 7 differs from hidden states learned for data from CK+ with expression variations shown in Figure 3. As a result, the global structure among hidden states, facial expression and poses of the model shown in Figure 6 (c) is different from the model shown in (a). The overall detection accuracies are shown in Table 2. Sample image results are shown in Figure 9.

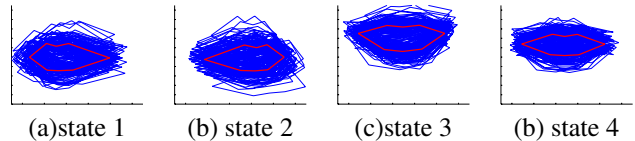

(a)state 1　　(b) state 2　　(c)state 3　　(b) state 4

Figure 7. Sampling results of the learned shape variations of mouth under different states for FERET database($P(X_{mouth}|Z_{mouth})$).

Table 2. Detection results on different databases.

|       | Eyebrow | Eye    | Nose   | Mouth   | Overall |
|-------|---------|--------|--------|---------|---------|
| MMI   | 5.8380  | 3.0120 | 4.9997 | 6.6119  | 5.0776  |
| FERET | 8.4713  | 3.1424 | 8.1405 | 8.8625  | 6.9011  |
| LFPW  | 8.4910  | 5.0231 | 9.1712 | 10.2489 | 8.0695  |
| ISL   | 9.5683  | 4.0264 | 6.8812 | 7.4248  | 6.7902  |

#### 4.3.3 Cross database facial feature detection

Table 3 shows facial feature detection results with cross database training and testing on CK+ and MMI databases. Compared to the within database facial feature detection results shown in Table 1 and 2, cross database facial feature detection achieves similar accuracy for the MMI database. But the error increases for the CK+ database. These results suggest that training data from the CK+ database has more general variations. Hence, the model trained on the CK+

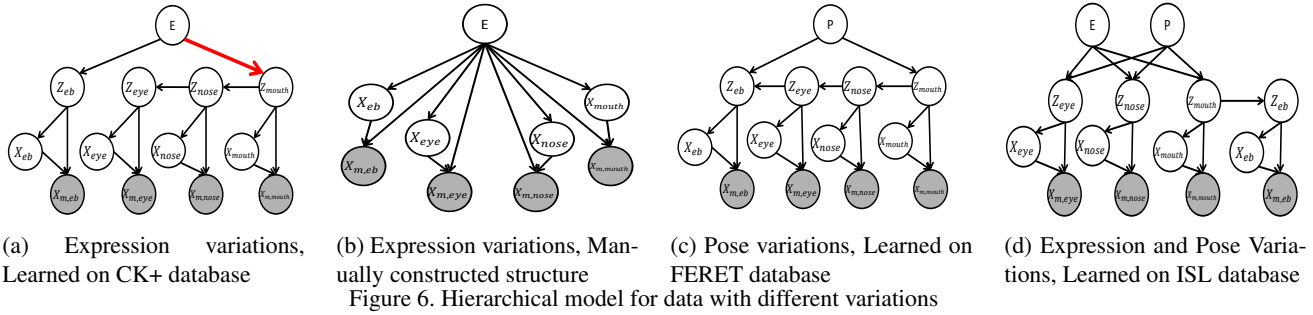

(a) Expression variations, Learned on CK+ database
(b) Expression variations, Manually constructed structure
(c) Pose variations, Learned on FERET database
(d) Expression and Pose Variations, Learned on ISL database

Figure 6. Hierarchical model for data with different variations

database can generalize to MMI. But, the converse is not true as shown in Table 3.

Table 3. Cross database facial feature detection.

| Training database | Testing database | Detection error |
|---|---|---|
| CK+ | MMI | 6.0369 |
| MMI | CK+ | 8.1584 |

### 4.4. Comparison with existing works

To compare our model with state-of-the-art works, we include the detection results using the RBM model [27], the BoRMaN model [25], the LEAR model [18], and the FPLL model [29]. We choose these models because they all utilize sophisticated face shape prior models constructed based on graphical models. For all the algorithms, we used the detection codes provided by their authors. For comparison, we used the shared points shown in Figure 8(a).

Detection results are shown in Table 4 and Figure 8 (b). Our detection results are consistently better than the RBM model [27], the BoRMaN model [25], the LEAR model [18], and the FPLL model [29]. For the ISL database, in which facial images are with significant facial expressions and poses, BoRMaN, LEAR and FPLL can not generate reasonable detection results. For the AFLW database, we also compare our algorithm with the Supervised Descent method in [28], which proposes a robust method to solve a nonlinear least squares problem for face alignment. As shown in Figure 8 (b), while our algorithm consistently outperforms RBM, BoRMan and FPLL, it is slightly worse than the Supervised Descent method in a small distance range.

Table 4. Comparison of algorithms on CK+, MMI, FERET, and ISL databases. Numbers with "*" represents the reported results in original papers.

|  | Expression | | Pose | Exp.& pose |
|---|---|---|---|---|
|  | CK+ | MMI | FERET | ISL |
| our algorithm | 4.05 | 5.04 | 6.91 | 6.77 |
| RBM [27] | 4.84* | 5.53* | 9.35 | 6.78* |
| BoRMaN [25] | 6.94 | 6.64 (4.65*) | 12.73 | fails |
| LEAR [18] | 9.27 | 6.24 (5.12*) | 9.24 | fails |
| FPLL [29] | 8.76 | 8.34 | 9.3 | fails |

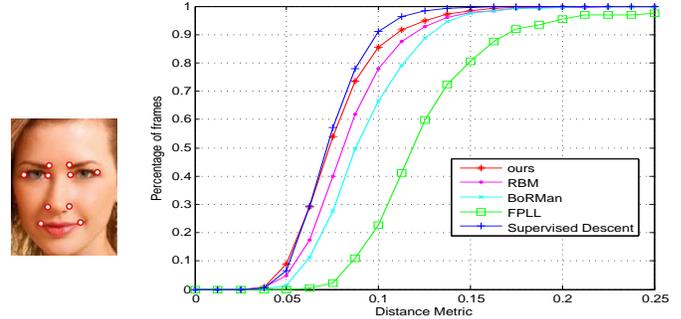

(a) shared points   (b) detection results

Figure 8. Comparison of algorithms on AFLW database

## 5. Conclusion

In this paper, we proposed a hierarchical probabilistic model that captures the face shape variations with varying facial expressions and poses and could infer the true facial feature locations given the measurements. There are two major benefits of this model. First, it implicitly models the local shape variation for each facial component. Second, it learns the joint relationship among facial components, the facial expression and the pose in the higher level by searching the optimal structure and parameterizations of the model. During testing, to infer true facial feature locations, the model combines the bottom-up information from local shape model and top-down constraints among facial components. Experimental results on benchmark databases demonstrate the effectiveness of the proposed hierarchical probabilistic model.

**Acknowledgements:** This work is supported in part by a grant from US Army Research office (W911NF-12-C-0017).

## References

[1] T. F. Cootes, G. J. Edwards, and C. J. Taylor. Active appearance models. *IEEE Transactions on Pattern Analysis and Machine Intelligence*, 23(6):681–685, 2001. 2

[2] T. F. Cootes, C. J. Taylor, D. H. Cooper, and J. Graham. Active shape models their training and application. *Computer Vision and Image Understanding*, 61(1):38–59, 1995. 2

[3] D. Cristinacce and T. Cootes. Automatic feature localisation with constrained local models. *Pattern Recognition*, 41(10):3054–3067, 2008. 2

[4] M. Dantone, J. Gall, G. Fanelli, and L. Van Gool. Real-time facial feature detection using conditional regression forests. In *Computer*

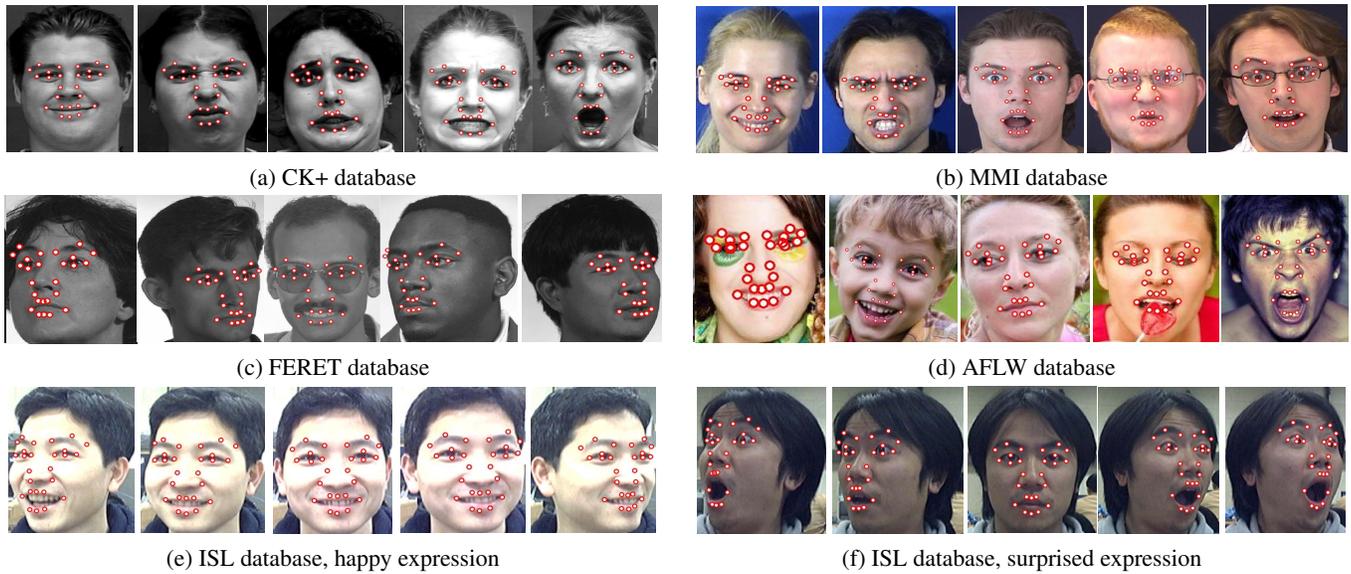

Figure 9. Facial feature detection results of sample images from different databases.

*Vision and Pattern Recognition (CVPR), 2012 IEEE Conference on*, pages 2578–2585, 2012. 2

[5] J. Daugman. Complete discrete 2-d gabor transforms by neural networks for image analysis and compression. *Acoustics, Speech and Signal Processing, IEEE Transactions on*, 36(7):1169–1179, 1988. 4

[6] C. P. de Campos, Z. Zeng, and Q. Ji. Structure learning of Bayesian networks using constraints. In *ICML '09: Proceedings of the 26th Annual International Conference on Machine Learning*, pages 113–120, New York, NY, USA, 2009. ACM. 5

[7] L. Ding and A. Martinez. Features versus context: An approach for precise and detailed detection and delineation of faces and facial features. *Pattern Analysis and Machine Intelligence, IEEE Transactions on*, 32(11):2022–2038, 2010. 2

[8] B. Fasel and J. Luettin. Automatic facial expression analysis: A survey. *PATTERN RECOGNITION*, 36(1):259–275, 1999. 1

[9] D. Fleet and A. Jepson. Computation of component image velocity from local phase information. *International Journal of Computer Vision*, 5(1):77–104, 1990. 4

[10] N. Friedman. Learning belief networks in the presence of missing values and hidden variables. In *Proceedings of the Fourteenth International Conference on Machine Learning*, pages 125–133. Morgan Kaufmann, 1997. 4

[11] R. Gross, I. Matthews, and S. Baker. Generic vs. person specific active appearance models. *Image and Vision Computing*, 23(11):1080–1093, November 2005. 2

[12] Y. Huang, Q. Liu, and D. Metaxas. A component-based framework for generalized face alignment. *Systems, Man, and Cybernetics, Part B: Cybernetics, IEEE Transactions on*, 41(1):287–298, 2011. 2

[13] T. Kanade, J. Cohn, and Y. Tian. Comprehensive database for facial expression analysis. In *Automatic Face and Gesture Recognition, 2000. Proceedings. Fourth IEEE International Conference on*, pages 46–53, 2000. 5

[14] M. Koestinger, P. Wohlhart, P. M. Roth, and H. Bischof. Annotated facial landmarks in the wild: A large-scale, real-world database for facial landmark localization. In *First IEEE International Workshop on Benchmarking Facial Image Analysis Technologies*, 2011. 5

[15] D. Koller and N. Friedman. *Probabilistic Graphical Models: Principles and Techniques*. MIT Press, 2009. 5

[16] V. Le, J. Brandt, Z. Lin, L. D. Bourdev, and T. S. Huang. Interactive facial feature localization. In *In procedding of European Conference on Computer Vision*, pages 679–692, 2012. 2

[17] P. Lucey, J. Cohn, T. Kanade, J. Saragih, Z. Ambadar, and I. Matthews. The extended cohn-kanade dataset (ck+): A complete dataset for action unit and emotion-specified expression. In *Computer Vision and Pattern Recognition Workshops (CVPRW), 2010 IEEE Computer Society Conference on*, pages 94–101, 2010. 5

[18] B. Martinez, M. F. Valstar, X. Binefa, and M. Pantic. Local evidence aggregation for regression-based facial point detection. *IEEE Transactions on Pattern Analysis and Machine Intelligence*, 35(5):1149–1163, 2013. 2, 7

[19] E. Murphy-Chutorian and M. Trivedi. Head pose estimation in computer vision: A survey. *Pattern Analysis and Machine Intelligence, IEEE Transactions on*, 31(4):607–626, 2009. 1

[20] M. Pantic, M. Valstar, R. Rademaker, and L. Maat. Web-based database for facial expression analysis. In *Multimedia and Expo, 2005. ICME 2005. IEEE International Conference on*, pages 5 pp.–, 2005. 5

[21] P. Phillips, H. Moon, P. Rauss, and S. Rizvi. The feret evaluation methodology for face-recognition algorithms. In *Computer Vision and Pattern Recognition, 1997. Proceedings., 1997 IEEE Computer Society Conference on*, pages 137–143, 1997. 5

[22] J. M. Saragih, S. Lucey, and J. F. Cohn. Deformable model fitting by regularized landmark mean-shift. *Int. J. Comput. Vision*, 91(2):200–215, Jan. 2011. 2

[23] Y. Tong, W. Liao, and Q. Ji. Isl multi-view facial expression database. http://www.ecse.rpi.edu/~cvrl/database/database.html. 5

[24] Y. Tong, Y. Wang, Z. Zhu, and Q. Ji. Robust facial feature tracking under varying face pose and facial expression. *Pattern Recognition*, 40:3195–3208, 2007. 2, 4

[25] M. Valstar, B. Martinez, X. Binefa, and M. Pantic. Facial point detection using boosted regression and graph models. In *Computer Vision and Pattern Recognition (CVPR), 2010 IEEE Conference on*, pages 2729–2736, 2010. 2, 7

[26] P. Viola and M. Jones. Rapid object detection using a boosted cascade of simple features. pages 511–518, 2001. 4

[27] Y. Wu, Z. Wang, and Q. Ji. Facial feature tracking under varying facial expressions and face poses based on restricted boltzmann machine. In *Proceedings of the International Conference on Computer Vision and Pattern Recognition*, pages 3452–3459, 2013. 2, 7

[28] X. Xiong and F. De la Torre Frade. Supervised descent method and its applications to face alignment. In *IEEE International Conference on Computer Vision and Pattern Recognition (CVPR)*, May 2013. 7

[29] X. Zhu and D. Ramanan. Face detection, pose estimation, and landmark localization in the wild. In *Computer Vision and Pattern Recognition (CVPR), 2012 IEEE Conference on*, pages 2879–2886, 2012. 2, 7